# DETECTING UNSAFE BEHAVIOR IN NEURAL NETWORK IMITATION POLICIES FOR CAREGIVING ROBOTICS

## A. TYTARENKO

**Abstract.** In this paper, the application of imitation learning in caregiving robotics is explored, aiming at addressing the increasing demand for automated assistance in caring for the elderly and disabled. Leveraging advancements in deep learning and control algorithms, the study focuses on training neural network policies using offline demonstrations. A key challenge addressed is the "Policy Stopping" problem, crucial for enhancing safety in imitation learning-based policies, particularly diffusion policies. Novel solutions proposed include ensemble predictors and adaptations of the normalizing flow-based algorithm for early anomaly detection. Comparative evaluations against anomaly detection methods like VAE and Tran-AD demonstrate superior performance on assistive robotics benchmarks. The paper concludes by discussing the further research in integrating safety models into policy training, crucial for the reliable deployment of neural network policies in caregiving robotics.

**Keywords:** assistive robotics, reinforcement learning, diffusion models, imitation learning, anomaly detection

## INTRODUCTION

In recent years the fields of robotics and AI attracted lots of interest. The advances in deep learning, robotics hardware, deep reinforcement learning, and imitation learning made it possible to solve complex control problems by training a neural network policy from mere hundreds of demonstrations.

In this paper, caregiving robotics is considered. Given the growing numbers of elderly and disabled people who need daily physical care [1, 2], the importance of automation rapidly increases. Caregiving (or assistive) robotics has a promise of addressing this problem, especially in the light of advances in control algorithms and hardware.

As in most human-robot interaction scenarios, one of the biggest concerns in caregiving control algorithms is safety. This concern is especially important with neural network-based policies, which lack interpretability and are known to become unstable on out-of-distribution data [3].

For the case of imitation learning, this problem is visualized on Fig 1.

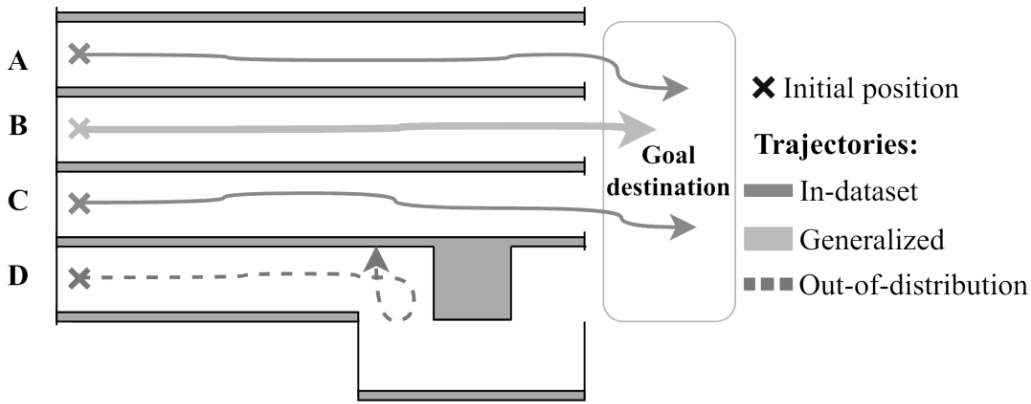

Fig 1. Out-of-distribution data may lead to failures of a policy. There are 4 episodes: A, B, C, D visualized as trajectories from an initial position marked as X to goal region. A and C are present in dataset. B is not present, but since it does not differ much from A and C, the algorithm is able to generalize. The episode D, however, is significantly different, and thus, a policy makes unexpected wrong decisions, failing the task.

The progress in the field is nevertheless vast. [4] proposes a method for robotic arm for assistive manipulation tasks. It is a learning-based system, capable of learning from demonstration, based on Dynamic Movement Primitives [5]. DMP is a vast framework that includes many instances. Although those methods give a potential for lifelong/incremental learning, they also rely careful modelling and are more difficult to implement and deploy.

Paper [6] introduced simulation software for assistive manipulation tasks, named AssistiveGym. It comes with multiple predefined tasks (feeding, drinking, arm manipulation, etc.) and robots (Jaco, PR1, etc.) to pick. For the study, this simulator is chosen for its versatility, simplicity, and speed. The simulator also comes with a Proximal Policy Optimization-based (PPO, [7, 8]) baseline. In this work, an imitation learning-based approach is used for training a neural network policy. Imitation learning [9, 10, 11] allows to avoid the necessity of learning from interaction, by instead leveraging the offline data (demonstrations) collected using an existing policy or via teleoperation.

The uncertainty estimation problem for Reinforcement Learning algorithms is studied in [12]. Although applied to a different task, the authors show that the uncertainty can be estimated using the log-likelihood and the variance of the model. The problem is, DDPMs in general, and Diffusion Policy specifically, is a generative model, for which calculating a likelihood for the generated plan is difficult [13], making the proposed approach hardly applicable for the considered problem. Other methods include [14, 15, 16, 17]

In the following sections, the "Policy Stopping problem" is studied and solutions are proposed. These solutions are compared to the application of out-of-box anomaly detection and uncertainty detection methods, proved to be successful in other domains. A system with a safety model and an imitation policy is developed and demonstrated. Lastly, the paper concludes with the discussion of the results and further research.

**PRELIMINARIES**

**Markov Decision Process** (MDP) is a collection $(S, A, r, T)$ with $S$ – state space, $A$ – action space, $r(s,a)$ – reward function and $T = P(s_{t+1} | s_t, a_t)$ – dynamics. In this paper, the reward is not assumed to be defined for full trajectories, classifying them as either "success" or "failure".

**Reinforcement Learning** (RL) algorithms optimize a policy $\pi$, which maximizes the expected total reward of the MDP:

$$\pi^* = \arg\max_\pi \mathrm{E}_{\tau \sim p(\cdot)} \sum_{t=0}^{\infty} \gamma r(s_t, a_t)$$

Where $\tau$ is a trajectory $(s_0, a_0, s_1, a_1, ... s_T)$ sampled by applying a policy $\pi$.

In offline setting (offline RL), an access to environment for collecting more interactions is assumed to be absent, and the whole training is conducted using only pre-collected demonstrations.

**Diffusion Policy**, is essentially a Denoising Diffusion Probabilistic Model (DDPM) which models a distribution $p(A|O)$, where $O$ is a subset of prior observations, and $A$ is a limited sequence of further actions, i.e. a short-horizon plan.

**Normalizing flow**-based methods [18] estimate the data likelihood explicitly, by using a reversible block of various kinds. A trained network maps the input data $x$ to latent space $Z$, such that the inverse mapping $f^{-1}(f(x))$ is trivially computable.

## METHOD

### Data collection

In this work, imitation learning techniques are used to train a neural network policy. Imitation learning methods as a rule require pre-recorded trajectories, e.g. a dataset with sequences of a form:

$$D = \{(s_0, a_0, s_1, a_1, ... s_T)_i, i = 1, ..., N\}$$

Here $N$ – is the number of trajectories and $T$ – is a length of a trajectory. For collecting the trajectories, two methods are used – teleoperation and online reinforcement learning algorithms.

Teleoperation is a fairly difficult task when it comes to robotic arm manipulation problems, especially in simulation. A keyboard-based teleoperation feature from the original AssistiveGym implementation is adapted for the task. The modified version is available via GitHub [19].

Online reinforcement learning algorithms allow training a policy neural network by interacting with an environment. They are usually way less sample-efficient, i.e. it takes much more data and training steps to learn a useful behaviour. Nevertheless, it is convenient in case of AssistiveGym, since some tasks are very difficult to teleoperate. Proximal Policy Optimization [7] algorithm is used, which is a well-established baseline Reinforcement Learning method, to collect useful trajectories for some of the tasks.

### Diffusion Policy for Assistive Robotics

Recent advances brought much more efficient imitation learning methods, such as Diffusion Policy [10] and Action Chunk Transformer [20]. Diffusion Policy, for instance, allows to train a relatively small neural network policy from up to 200-300 demonstrations in some cases [10].

Diffusion Policy fits a network capable of producing a plan of actions $A$ from $P(A|S)$ without explicitly learning it. More precisely,

$$S = (s_{k-T_O}, ..., s_k),$$
$$A = (a_{k-T_O}, ..., a_k, ..., a_{k+T_A}),$$

where $k$ is a current time step, $T_A$ – action plan horizon, and $T_O$ – state (observation) horizon. In this work, $S$ is a concatenation of previous states, each of which is represented as a vector of real numbers, i.e. $s_t \in \Re^{N_S}$. In the current study, $N_S$ is a relatively small number (<100), although the method allows working with larger-dimensional state spaces. This description also applies to the action plan $A$: $a_t \in \Re^{N_a}$.

The problem, however, is that it is difficult to compute a likelihood of a sample given a model only, which means that there is only a short-horizon plan **A** without any additional information.

Although the method is known to be sample-efficient, it still highly depends on the quality of the dataset, i.e. state-space coverage, trajectory optimality, etc. See Fig 1.

Therefore, there are almost none guarantees that a deployed robotic policy won't fail in unexpected ways, potentially damaging the hardware. Moreover, since the Caregiving Robotics deals with human-robot interaction, this may make the robot dangerous to a human, which is a critical in this domain.

**Policy stopping problem**

In this study, approaches to the "Policy Stopping" problem are proposed and compared. In it, an algorithm must decide whether a policy execution must be stopped immediately. This problem can be also viewed as an early anomaly detection problem. However, there is one important difference. The stopping algorithm must be trained on offline data, generated by a behavioural policy (a human demonstrator, a scripted policy, arbitrary neural network policy, or a mix), but tested on a data, generated by a different policy trained on that data (e.g. imitation learning algorithm).

The key difference from traditional unsupervised anomaly detection is that an algorithm is conditioned on a dataset, generated by a distribution different from the test one. Therefore, such algorithm must balance the similarity of test trajectory and train trajectories, distinguishing between a good plan executed successfully but in unusual way and a bad plan that ends up in failure.

**State-prediction approach**

The first approach considered is inspired by MBPO [14] and widely used in Reinforcement Learning algorithms for different purposes [21, 22, 23]. This approach uses a "disagreement" of an ensemble of next state prediction neural networks. The idea is that the next state prediction will be accurate and won't vary much between networks in the ensemble if the input is in-distribution (familiar to the model). At the same time, a state-action pair may not be known. The reason may be that it was not present in a dataset or that a dataset does not contain enough data for a predictor to generalize successfully to this state-action pair. Then, the next state predictors will "disagree", which can be measured as a variance of some kind.

Based on that principle, a network is trained, approximating a function

$$f_\phi(S \mid A) = S',$$

which predicts a vector of $T_{out}$ future states.

For training, inputs and outputs are sampled from a collection of trajectories and a neural network is fit in a simple supervised way, minimizing the MSE (Mean Squared Error) objective:

$$L_{MSE}(S, A, S', \phi) = \| f_\phi(S, A) - S' \|_2^2$$

Sampling is executed in a following way:

$$(s_0, \ldots, s_T) \sim D_{demon},$$
$$k \sim \{k, \ldots, T-k\},$$
$$S = (s_{k-T_{in}}, \ldots, s_k),$$
$$A = (a_{k-T_{in}}, \ldots, a_k),$$
$$S = (s_{k+1}, \ldots, s_{k+T_{out}}),$$

An ensemble of $K$ models is trained, by initializing and fitting them independently on the same data. For estimating the level of uncertainty, a standard deviation between state predictions is computed by the following formula:

$$U^l(S, A) = \frac{1}{K-1} \sqrt{\sum_{i=1..K} \left[ f_{\phi_i}(S, A)^l - \frac{1}{K} \sum_{j=1..K} f_{\phi_j}(S, A)^l \right]^2},$$
$$U_{ESP}(S, A) = \sum_{l=1..|S|N_S} U^l(S, A),$$

where $\phi_i$ ($i = 1..K$) – parameters of neural networks in an ensemble.

After computing the uncertainty level $U_{ESP}$, an algorithm compares it to a manually tuned threshold and returns a decision for whether to stop an episode or not. See the Pseudocode 1 for details.

Pseudocode 1. Training an ensemble state prediction model.
1. Input: Dataset $D_{demon}$.
2. Initialize hyperparameters $T_{in}, T_{out}, K$.
3. For $N_E$ epochs:
4. For $j = 1..K$:
5. Sample $S, A, S'$.
6. Compute the MSE loss $L_{MSE}(S, A, S', \phi)$.
7. Compute the gradients w.r.t. $\phi_j$, update the weights $\phi_j$.
8. End for.
9. End for.
10. Return: $\phi_1..\phi_K$.

The considered approach follows [14] with a difference that the input to the state prediction function is not necessarily a single state-action pair, but a chunk, or an entire sub-trajectory. Although excessive due to the assumed Markovianess of the MDP, this allows to incorporate correlations between earlier states and decisions made by an agent, such the resulting neural network ensemble shall disagree when there are longer-term non-immediate anomalies in entire sub-trajectories and not only a single state-action pair.

In other words, single state-action version computes $U_{ESP}(s, a)$, while the proposed one computes $U_{ESP}(S, A)$.

In this study, a simple MLP (multi-layered perceptron) architecture is used for a single state-action version, and a CNN (convolutional neural network) is used for the proposed sub-trajectory version.

**Adapting anomaly detection methods based on normalizing flows**

A promising approach in unsupervised anomaly detection is normalizing flows.

In this paper, a method named MVT Flow [18] is considered. MVT Flow is designed for unsupervised anomaly detection in time series in a robotics domain. Using a convolutional neural network as a backbone, it is trained to estimate the likelihood of normal data. The anomaly score is then computed as a loss function of a test data w.r.t. the trained model.

MVT Flow can't be successfully applied to the presented problem out of box. Although [18] provides a method for credit assignment of elements of the series, it still requires a network to process the entire time series first. Thus, to adapt MVT Flow to early anomaly detection setting the following modification is proposed.

**Masking augmentation and sample weighting**

The anomaly detection method MVT-Flow is (i) unsupervised and (ii) assigns an anomaly score to the entire input sequence. Therefore, applying it to a not finished sequence may be problematic. The neural net directly maximizes the likelihood of training data, so a previously unseen sequence will get a low likelihood score and will be considered anomaly.

First, unfinished sequences are added to the training data, by randomly choosing a sub-episode length and removing all following elements from the episode. The problem, however, is that the actual abnormal trajectory may start as a normal one with only minor differences. Resulting model does not distinguish between a beginning of a normal trajectory and a fully normal trajectory, where clearly the likelihood should be different.

So, second, a sample weighting is introduced to compensate for that effect:

$$w = \max\left(\sqrt{\frac{K - K_{min}}{K_{max} - K_{min}}}, w_0\right),$$

where $K, K_{min}, K_{max}$ are respectively a sub-episode length, a minimum sub-episode length and a full episode length.

Intuitively, the ratio under the square root is a value which is 0 when the sub-episode is minimal and 1 when the sub-episode is full. The square root is applied to smooth the weights, making the difference between the full episode and minimal one smaller.

**Full algorithm**

Pseudocode 2. Training an early-detection MVT-Flow model.
1. Input: Dataset $D_{demon}$.
2. Initialize hyperparameters $\gamma, N_E, K_{min}, K_{max}, w_0$.
3. For $N_E$ epochs:
4. Sample $S_r, A_r \sim D_{demon}$.

5. Sample random sub-episode length $K \sim \{K_{min},...,K_{max}\}$.
6. Compute masked data $S, A$:
$$S^i = I_{i<K}, i = 1...|S|$$
$$A^i = I_{i<K}, i = 1...|A|$$
7. Compute the MVT-Flow loss $L_{MVT}(S, A, \theta)$.
8. Compute the sample weight: $w = \max\left(\sqrt{\dfrac{K - K_{min}}{K_{max} - K_{min}}}, w_0\right),$.
9. Update weights: $\theta := \theta - \gamma w \nabla_\theta L_{MVT}(S, A, \theta)$.
10. End for.
11. End for.
12. Return: weights $\theta$.

**EXPERIMENTAL VALIDATION**

In this section, the results of the study on several benchmarks of Caregiving Robotics are provided. All benchmarks are conducted using environments from the modified version of the AssistiveGym suit, available via GitHub [19].

A simulated Jaco robotic arm is used, the following assistive tasks are considered: Assistive Feeding (250 teleoperation deomnstrations), Assistive Bed Bathing (1000, PPO), Arm Manipulation (1000, PPO), and Scratch Itch (1000, PPO).

First, a policy network is trained using a diffusion policy algorithm on each collected dataset with trajectories.

Next, on each dataset, the models of weighted-masked (WM) MVT-Flow, original MVT-Flow, ensemble state predictors (single state-action and sub-trajectory based), Variational Autoencoder (VAE) and Tran-AD.

Weighted-masked MVT-Flow is trained with $K_{min} = 1$, $K_{max} = 200$, $w_0 = 0.1$, $\gamma = 8 \cdot 10^{-4}$, $N_E = 85$. Other hyperparameters are kept in sync with [18].

Ensemble state predictors with $K = 5$. For single state-action version take $T_{in} = T_{out} = 1$, and for sub-trajectory based, take $T_{in} = t$, $T_{out} = 1$. Here $t$ means that all observations and actions observed up to a moment $t$ are considered. Single state-action predictor is applied sequentially, and a maximum uncertainty score is taken as a resulting anomaly score.

Variational Autoencoder has a small CNN backbone and *KL* penalty is set to 1. The anomaly score is set to the value of reconstruction loss of the input sub-episode.

For Tran-AD the window size is set to 20. For evaluation, a Tran-AD network is inferred on all windows contained within the sub-episode and the resulting anomaly score is set to maximum anomaly score of every window.

Every other hyperparameter remains unchanged from the original papers.

To evaluate the quality of the proposed models, two kinds of metrics are reported: AUROC and FPR@TPR95. The former one is defined as an area under the Receiver Operating Characteristic curve. The later one is defined as the False Positive Rate on a threshold corresponding to 0.95 True Positive Rate. Both are common metrics in anomaly detection literature [24].

However, since the goal is to evaluate the early anomaly detection property, the metrics are reported for partial trajectories of various maximum lengths, namely 10%, 20%, 30%, 50%, 75%, and 100% of the maximum episode length. Better metrics on smaller percentages correspond to better earlier detection ability of an evaluated method.

Tables 1-5 contain metrics reported when evaluated of each assistive environment datasets. Note, that for AUROC larger is better, while for FPR@TPR95 lower is better.

| Method | Metric | 10% | 20% | 30% | 50% | 75% | 100% |
|---|---|---|---|---|---|---|---|
| Single SP | FPR@TPR95 | 0.81 | 0.73 | 0.77 | 0.76 | 0.58 | 0.34 |
|  | AUROC | **0.79** | 0.80 | 0.79 | 0.81 | 0.90 | 0.92 |
| VAE | FPR@TPR95 | 0.70 | 0.76 | 0.73 | 0.45 | 0.18 | **0.001** |
|  | AUROC | 0.70 | 0.71 | 0.77 | 0.86 | 0.96 | **1.00** |
| MVT-Flow | FPR@TPR95 | 0.72 | 0.37 | 0.55 | **0.23** | **0.06** | 0.02 |
|  | AUROC | 0.70 | 0.80 | 0.80 | **0.94** | **0.98** | 0.99 |
| Tran-AD | FPR@TPR95 | 0.74 | 0.74 | 0.73 | 0.63 | 0.64 | 0.40 |
|  | AUROC | 0.65 | 0.70 | 0.69 | 0.79 | 0.89 | 0.92 |
| Sub-trajectory SP* | FPR@TPR95 | **0.51** | **0.63** | 0.60 | 0.33 | **0.07** | **0.001** |
|  | AUROC | **0.79** | **0.83** | 0.83 | 0.92 | 0.97 | **1.00** |
| WM MVT-Flow* | FPR@TPR95 | 0.64 | **0.61** | **0.50** | 0.21 | **0.06** | **0.001** |
|  | AUROC | 0.77 | **0.83** | **0.84** | **0.94** | **0.98** | **1.00** |

Table 1. Evaluation on Assistive Feeding

Assistive feeding is a simpler task, so most normal trajectories have a relatively short length. Therefore, it is expected that a good method gets maximum score on 100% of the environment length.

| Method | Metric | 10% | 20% | 30% | 50% | 75% | 100% |
|---|---|---|---|---|---|---|---|
| Single SP | FPR@TPR95 | 0.82 | 0.80 | 0.77 | 0.82 | 0.40 | 0.20 |
|  | AUROC | 0.73 | 0.72 | 0.79 | 0.78 | 0.89 | 0.95 |
| VAE | FPR@TPR95 | 0.82 | 0.80 | 0.73 | 0.37 | 0.26 | **0.02** |
|  | AUROC | 0.82 | 0.80 | 0.77 | 0.87 | 0.92 | **0.99** |
| MVT-Flow | FPR@TPR95 | 0.84 | 0.76 | 0.43 | **0.16** | **0.08** | **0.01** |
|  | AUROC | 0.72 | 0.75 | 0.88 | **0.95** | **0.97** | **0.99** |
| Tran-AD | FPR@TPR95 | 0.97 | 0.96 | 0.97 | 0.90 | 0.83 | 0.78 |
|  | AUROC | 0.43 | 0.43 | 0.45 | 0.51 | 0.57 | 0.65 |
| Sub-trajectory SP* | FPR@TPR95 | 0.84 | 0.80 | 0.85 | 0.41 | **0.10** | **0.02** |
|  | AUROC | 0.72 | 0.74 | 0.68 | 0.88 | **0.96** | **0.99** |
| WM MVT-Flow* | FPR@TPR95 | **0.72** | **0.71** | **0.67** | **0.16** | **0.07** | **0.03** |
|  | AUROC | **0.88** | **0.88** | **0.89** | **0.96** | **0.98** | **0.99** |

Table 2. Evaluation on Arm Manipulation

| Method | Metric | 10% | 20% | 30% | 50% | 75% | 100% |
|---|---|---|---|---|---|---|---|
| Single SP | FPR@TPR95 | 0.88 | 0.91 | 0.94 | 0.95 | 0.90 | 0.87 |
|  | AUROC | **0.79** | 0.65 | 0.66 | 0.64 | 0.66 | 0.67 |
| VAE | FPR@TPR95 | **0.84** | 0.79 | 0.79 | 0.76 | 0.54 | 0.85 |
|  | AUROC | 0.68 | 0.63 | 0.63 | 0.70 | 0.81 | **0.97** |
| MVT-Flow | FPR@TPR95 | 1.00 | 0.83 | 0.83 | 0.55 | 0.44 | 0.28 |
|  | AUROC | 0.40 | 0.66 | 0.71 | 0.77 | 0.82 | 0.82 |
| Tran-AD | FPR@TPR95 | 1.00 | 0.88 | 1.00 | 0.94 | 0.89 | 0.89 |
|  | AUROC | 0.50 | 0.53 | 0.51 | 0.51 | 0.52 | 0.54 |
| Sub-trajectory SP* | FPR@TPR95 | 0.88 | 0.87 | 0.80 | 0.82 | 0.72 | **0.001** |
|  | AUROC | **0.77** | **0.74** | **0.74** | 0.66 | 0.71 | **1.00** |
| WM MVT-Flow* | FPR@TPR95 | 0.87 | **0.69** | **0.67** | **0.50** | **0.40** | 0.22 |
|  | AUROC | 0.72 | **0.77** | **0.77** | **0.81** | **0.86** | 0.94 |

Table 3. Evaluation Assistive Bed Bathing

Bed bathing dataset is challenging due to the low success rate of the demonstration policy. Therefore, the distribution of input trajectories may not cover most scenarios, limiting an imitation learning policy's performance.

| Method | Metric | 10% | 20% | 30% | 50% | 75% | 100% |
|---|---|---|---|---|---|---|---|
| Single SP | FPR@TPR95 | 0.84 | 0.89 | 0.79 | 0.73 | 0.70 | 0.56 |
|  | AUROC | 0.60 | 0.63 | 0.66 | 0.69 | 0.80 | 0.82 |
| VAE | FPR@TPR95 | 0.95 | 0.89 | 0.84 | 0.77 | **0.29** | 0.17 |
|  | AUROC | 0.60 | 0.61 | 0.66 | 0.75 | **0.88** | **0.92** |
| MVT-Flow | FPR@TPR95 | 0.85 | 0.89 | 0.84 | **0.67** | 0.45 | 0.30 |
|  | AUROC | 0.56 | 0.65 | 0.70 | 0.75 | 0.84 | **0.90** |
| Tran-AD | FPR@TPR95 | **0.72** | **0.60** | **0.70** | 0.81 | 0.67 | 0.55 |
|  | AUROC | **0.77** | **0.79** | 0.75 | 0.71 | 0.79 | 0.83 |
| Sub-trajectory SP* | FPR@TPR95 | 0.88 | 0.84 | 0.82 | **0.68** | 0.38 | **0.07** |
|  | AUROC | 0.56 | 0.60 | **0.77** | **0.80** | 0.87 | **0.93** |
| WM MVT-Flow* | FPR@TPR95 | 0.81 | 0.83 | 0.84 | **0.65** | 0.41 | 0.30 |
|  | AUROC | 0.74 | **0.78** | **0.79** | 0.79 | 0.86 | **0.91** |

Table 4. Scratch Itch

From Tables 1-4, on can make the following observations.

First, Weighted-Masked MVT-Flow consistently outperforms raw MVT-Flow. For higher % of maximum length the raw version usually performs on par with the proposed modification, which is expected. This is because anomalous episodes take 100% of maximum length time, while most normal episodes are up to 50-75% of time.

Second, Sub-trajectory SP performs on par with WM MVT-Flow on simpler environments, such as Feeding. It also outperforms single step predictors, especially on larger time periods.

Tran-AD models perform the worst on most datasets due to its windowed inputs. The only exception is Scratch Itch (Table 4). It is hypothesized that the reason for this is the smaller-scale nature of anomalies in the test trajectories.

Now, a demonstration of a system with a diffusion policy deployed with a safety model is provided (see Fig 2). In practice, a set of thresholds for each time period is selected, since the anomaly score for applied methods is non-decreasing.

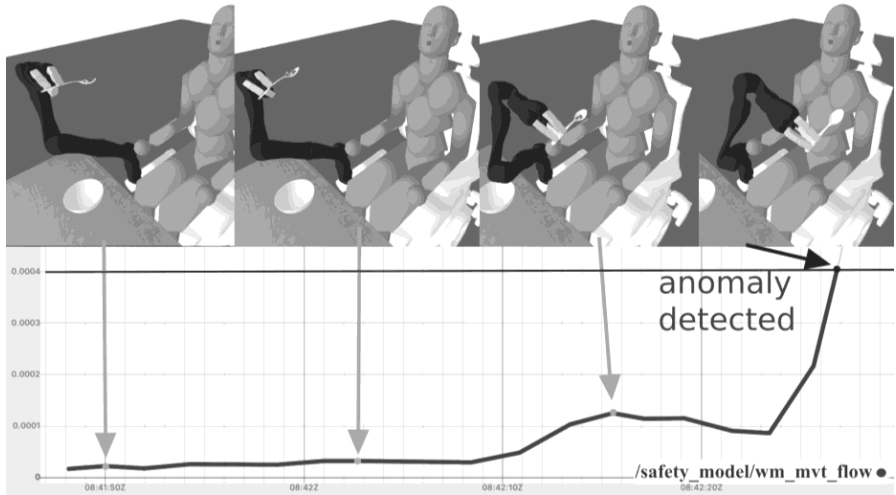

Fig 2. Demonstration of the proposed approach on the Assistive Feeding environment. The lower part of the diagram shows a plot of the anomaly score (normalized), and arrows matching the upper images with corresponding time steps. Most of the time, the score is low, since the arm performs usual moves. The end of the plot shows a spike in anomaly score, resulting in the system halt. The anomaly is that the arm drops food and spins itself in unusual way. In the remaining of this episode, the arm would twist itself dangerously, potentially damaging hardware.

**CONCLUSION**

In this paper, a challenging "Policy Stopping problem" is introduced and studied. This problem is important for improving safety of imitation learning-based neural network policies, specifically diffusion policies.

The solutions specific to the introduced problem are proposed: ensemble of sub-trajectory-based state predictors and a modification of a recent MVT-Flow algorithm for early anomaly detection.

The algorithms are evaluated and compared against ablated original unmodified versions and known anomaly detection approaches, such as VAE and Tran-AD. The proposed solutions are shown to be more suitable for the introduced problem and tend to outperform other methods on assistive robotics benchmarks. For the evaluation of early-detection capabilities the usual metrics have been adapted.

Lastly, a system with a safety model and an imitation policy is developed and demonstrated.

The interesting future work directions include integration of the proposed safety models to training of imitation policies (e.g. [21]), safe data collection for model finetuning, and adaptation of safety models to vision-based tasks. This may bring the safe and robust deployment of neural network policies, so important for caregiving robotics domain.